\documentclass[preprint]{article}
\usepackage{neurips_2025}
\usepackage[utf8]{inputenc} 
\usepackage[T1]{fontenc}    
\usepackage{hyperref}       
\usepackage{url}            
\usepackage{booktabs}       
\usepackage{amsfonts}       
\usepackage{nicefrac}       
\usepackage{microtype}      
\usepackage{xcolor} 
\usepackage{algorithm}
\usepackage{algorithmic}
\usepackage{amsmath}
\usepackage{graphicx}
\usepackage{amssymb}
\usepackage{subcaption}
\usepackage{booktabs}
\usepackage{natbib} 
\usepackage{bm} 
\usepackage{tikz}
\usepackage{longtable}
\usepackage{array}
\usetikzlibrary{shapes.geometric, arrows, positioning}
\usepackage{indentfirst}

\usepackage{enumitem}

\begin{document}

\title{Uncertainty Quantification in Probabilistic Machine Learning Models: Theory, Methods, and Insights
}


\author{
\textbf{Marzieh Ajirak}$^{1}$\thanks{Work done while at Stony Brook University.} \quad
Anand Ravishankar$^{2}$ \quad
Petar M. Djuri\'c$^{2}$ \\
$^{1}$Weill Cornell Medicine, Cornell University, New York, NY \\
$^{2}$Department of Electrical and Computer Engineering, Stony Brook University, Stony Brook, NY \\
\texttt{maa4083@med.cornell.edu} \quad
\texttt{\{anand.ravishankar, petar.djuric\}@stonybrook.edu}
}

\maketitle

\begin{abstract}
Uncertainty Quantification (UQ) is essential in probabilistic machine learning models, particularly for assessing the reliability of predictions. In this paper, we present a systematic framework for estimating both epistemic and aleatoric uncertainty in probabilistic models. We focus on Gaussian Process Latent Variable Models and employ scalable Random Fourier Features-based Gaussian Processes to approximate predictive distributions efficiently. We derive a theoretical formulation for UQ, propose a Monte Carlo sampling-based estimation method, and conduct experiments to evaluate the impact of uncertainty estimation. Our results provide insights into the sources of predictive uncertainty and illustrate the effectiveness of our approach in quantifying the confidence in the predictions.
\end{abstract}

\section{Introduction}
Uncertainty Quantification (UQ) plays a fundamental role in probabilistic machine learning (ML). ML models are increasingly used in applications that require making decisions and where reliable  predictions are essential. This is especially important in fields that demand reliable choices under uncertainty, such as medical diagnosis 
\cite{abdar2022need}
autonomous systems 
\cite{cheng2023machine}, 
and scientific modeling in general 
\cite{ghanem2017handbook}. However, most traditional ML models provide only point predictions without assessing their confidence. This, in turn, makes it difficult to quantify the reliability of the results 
\cite{papamarkou2024position}.

To address this limitation, probabilistic ML models provide a foundational framework for representing uncertainty \cite{murphy2012machine}. By explicitly modeling uncertainty \cite{wang2025aleatoric}, these models not only improve reliability in high-risk applications, but also promote better generalization by distinguishing between known and unknown regions in the input space. Moreover, uncertainty estimates increase model interpretability by clarifying the sources of uncertainty and revealing potential limitations and biases, which emphasizes the role of UQ in characterizing and managing the uncertainty inherent in models and their predictions while providing valuable insights into reliability and confidence calibration.
Given the broad impact of UQ, improving uncertainty estimation remains a pressing challenge.

In probabilistic ML models, uncertainty can be categorized into two main types:
\begin{enumerate}
    \item {Epistemic uncertainty} (\(\mathcal{E}\)): Also known as model uncertainty, this type of uncertainty arises due to a lack of knowledge about the model, which, in turn, results from insufficient training data. It reflects the model's confidence in its parameters and structure.
    \item {Aleatoric uncertainty} (\(\mathcal{A}\)): This uncertainty, often called data uncertainty, originates from intrinsic noise in the data-generating process and represents variability that remains irreducible, even with unlimited data.
\end{enumerate}
The total predictive uncertainty in a model is given by
\begin{align}
    \mathcal{T} &= \mathcal{E} + \mathcal{A},
\end{align}
where \(\mathcal{T}\) represents the total uncertainty in the model’s prediction.

In this paper, we focus on developing a principled framework for estimating both epistemic and aleatoric uncertainty in probabilistic ML models. Since probabilistic ML models comprise a broad range of approaches, we focus on making the problem more tractable by studying the UQ of predictions made by Gaussian process latent variable models (GPLVMs) \cite{titsias2010bayesian}.  
{Specifically, we explore how uncertainty can be efficiently quantified using scalable approximations, such as Random Fourier Features (RFF)-based Gaussian processes (GPs), which we use to construct GPLVMs.}

Our contributions include:
\begin{itemize}
    \item A systematic formulation of epistemic and aleatoric uncertainty estimation.
    \item A scalable approach to UQ using RFF-based GPs.
    \item Methodology for computing epistemic and aleatoric uncertainties, along with insights gained from experimental results.
\end{itemize}
The paper is organized as follows. First, we provide background on probabilistic ML models in Section \ref{sec:back}, followed by a brief overview of GPLVMs in Section \ref{sec:GPLVM}. In Section \ref{sec:theory}, we present the theoretical foundation for evaluating epistemic and aleatoric uncertainties, while in Section \ref{sec:methods} we provide details of the methods used for these evaluations. Section \ref{sec:insights} discusses experimental results and insights gained from various experiments. Finally, Section \ref{sec:conc} concludes with final remarks.

\section{Background}
\label{sec:back}
A probabilistic ML model learns a predictive distribution over outputs \(\bm y_n\)
given an input $\bm x_n$. The general formulation takes the following form:
\begin{align}
\label{eq:predd}
    p(\bm y_n | \bm x_n, \mathcal{D}) = \int p(\bm y_n | \bm x_n, \bm\theta) p(\bm \theta | \mathcal{D}) {\rm d}\bm\theta,
\end{align}
where \( \mathcal{D} = \{\bm x_n, \bm y_n\}_{n=1}^{N} \) is the training dataset, \( p(\bm\theta | \mathcal{D}) \) is the posterior distribution over the model parameters, and 
     \( p(\bm y_n | \bm x_n, \bm \theta) \) represents the likelihood function. 

In many cases, ML models approximate this likelihood function using a deterministic mapping from inputs to outputs. In such models, a function \( f(\bm x_n, \bm \theta) \) produces a single output estimate,  
\begin{align}
    \widehat{\bm y}_n = f(\bm x_n, \widehat{\boldsymbol{\theta}}),
\end{align}
where $\widehat{\bm\theta}$ is some estimate of $\bm\theta$ obtained from the training data ${\cal D}$.
However, probabilistic models extend this by modeling the entire predictive distribution rather than just a point estimate as given by \eqref{eq:predd}.

We now describe how this predictive distribution quantifies the uncertainty of the prediction through its total covariance. Let the predictive distribution be given as in \eqref{eq:predd}. Then, we can express the predictive mean as follows:
\begin{align}
\mathbb{E}_{         p(\bm{y}_n | \bm{x}_n, \mathcal{D})} [\bm{y}_n] = \mathbb{E}_{p(\boldsymbol{\theta} | \mathcal{D})} \big[ \mathbb{E}_{p(\bm{y}_n | \bm{x}_n, \boldsymbol{\theta})} [\bm{y}_n] \big].
\end{align}
To find the total predictive covariance, we apply the law of total covariance and obtain
\begin{align}
\text{Cov}_{p(\bm{y}_n | \bm{x}_n, \mathcal{D})} [\bm{y}_n] &= 
\text{Cov}_{p(\boldsymbol{\theta} | \mathcal{D})} \big[ \mathbb{E}_{p(\bm{y}_n | \bm{x}_n, \boldsymbol{\theta})} [\bm{y}_n] \big]\notag\\
& +
\mathbb{E}_{p(\boldsymbol{\theta} | \mathcal{D})} \big[ \text{Cov}_{p(\bm{y}_n | \bm{x}_n, \boldsymbol{\theta})} [\bm{y}_n] \big].
\end{align}
This decomposition separates epistemic uncertainty (first term) from aleatoric uncertainty (second term), i.e.,
    \begin{align}
        \mathcal{E}  &= 
\text{Cov}_{p(\boldsymbol{\theta} | \mathcal{D})} \big[ \mathbb{E}_{p(\bm{y}_n | \bm{x}_n, \boldsymbol{\theta})} [\bm{y}_n] \big],\\
        \mathcal{A} &= \mathbb{E}_{p(\boldsymbol{\theta} | \mathcal{D})} \big[ \text{Cov}_{p(\bm{y}_n | \bm{x}_n, \boldsymbol{\theta})} [\bm{y}_n] \big].
    \end{align}
In words, the epistemic uncertainty measures how much the expected prediction varies due to uncertainty in the parameters $\bm\theta$. It is obtained by first computing the expectation over the predictive distribution given the parameters and then computing the covariance of this expectation across different model parameters sampled from the posterior of the parameters. The aleatoric uncertainty is the expected predictive covariance due to data noise under the model's posterior.

\section{GPLVMs and inference with GPLVMs }
\label{sec:GPLVM}
We now briefly introduce GPLVMs with a focus on the uncertainties in their predictions. 
GPLVMs are powerful tools for modeling high-dimensional data by learning low-dimensional latent representations 
\cite{titsias2010bayesian}.
They establish a connection between a low-dimensional latent space and a high-dimensional observed space through GPs. Specifically, GPLVMs model data $\bm Y\in\mathbb{R}^{N\times d_y}$ in a high-dimensional space $\mathcal{Y}$, where $\mathcal{Y} = \mathbb{R}^{d_y}$, by assuming that these data are generated from latent variables $\bm X$ in a lower-dimensional space, $\mathcal{X}$, with $\bm X\in\mathbb{R}^{N\times d_x}$ and $\mathcal{X} = \mathbb{R}^{d_x}$, where $d_x \ll d_y$. Here, $N$ represents the number of paired observations between $\mathcal{X}$ and $\mathcal{Y}$. 

More specifically, each observed vector $\boldsymbol{y}_n\in\mathbb{R}^{d_y}$, $n=1, 2, \ldots, N$ is  generated from a corresponding latent vector  $\boldsymbol{x}_n\in\mathbb{R}^{d_x}$, where the mapping $\boldsymbol{x}_n\mapsto \boldsymbol{y}_n$ is modeled using $d_y$ GPs \cite{rasmussen2006gaussian}. In other words, for a given latent variable $\bm x_n$, the observed variable $\bm y_n$ is drawn according to 
\begin{align}
    \bm y_n&\sim {\cal N}(\bm f,\bm \Sigma_y),
\end{align}
where $\bm \Sigma_y={\rm diag}\{\sigma_1^2, \sigma_2^2,\ldots, \sigma_{d_y}^2\}$,  and $\bm f$ is a vector of $d_y$ functions, i.e.,  $\bm f(\bm x_n)=[f_1(\bm x_n), f_2(\bm x_n),\ldots, f_{d_y}(\bm x_n)]^\top$, each generated from a GP prior,
\begin{align}
f_d(\cdot)&\sim {\cal GP}(0,\kappa_d), \;\;\;d=1, 2, \ldots, d_y.
\end{align}







Now we provide a brief overview of inference with GPLVMs. Given the observed (training) data $\bm Y\in\mathbb{R}^{N\times d_y}$,  
the main objective during training is to learn how to draw samples from the posterior distribution,
\color{black}
\begin{align}
\label{eq:neweq1}
p(\boldsymbol{X},\boldsymbol{\theta}, \boldsymbol{\Sigma}_y|\boldsymbol{Y}) &\propto p(\boldsymbol{Y}|\boldsymbol{X},\boldsymbol{\theta},\boldsymbol{\Sigma}_y)p(\boldsymbol{X})p(\boldsymbol{\theta})p(\boldsymbol{\Sigma}_y),
\end{align}
where 
$\boldsymbol{\theta}$ represents the model parameters, 
$p(\boldsymbol{Y}|\boldsymbol{X},\boldsymbol{\theta}, \bf\Sigma_y)$ 
is the likelihood function, where we treat 
$\boldsymbol{Y}$ as observed variables and view $\boldsymbol{X},\boldsymbol{\theta},$ and $\bm\Sigma_y$ as the unknowns to be inferred, and 
$p(\boldsymbol{X})$,  $p(\boldsymbol{\theta})$ and $p(\boldsymbol{\Sigma})$ are the respective priors of  
$\boldsymbol{X}$, $\boldsymbol{\theta}$, and $\boldsymbol{\Sigma}_y$. Samples from the posterior can be drawn using methods such as Hamiltonian Monte Carlo (HMC) or other Markov Chain Monte Carlo (MCMC) techniques.

For testing, let $\bm y_*=[\bm y_{o,*}^\top, \bm y _{u,*}^\top]^\top$, where $\bm y_{o,*}$ and $\bm y_{u,*}$ represent the observed and unobserved parts of $\bm y_*$. The goal is now to determine the predictive distribution of $\bm y_{u,*}$ given $\bm y_{o,*}$, $p(\bm y_{u,*}|\bm y_{o,*},$ $\bm Y)$, using the learned posterior of the GPLVM in \eqref{eq:neweq1}. 
We achieve this in three steps:. 
\begin{enumerate}
\item First we infer the posterior distribution of the latent variable $\bm x_*$ that corresponds to the partially observed test point $\bm y_{o,*}$. This posterior is given by  
\begin{align}
\label{eq:normals}
    p(\boldsymbol{x}_*|\boldsymbol{y}_{o,*},\boldsymbol{Y})&\propto p(\boldsymbol{x}_*)\int  p(\boldsymbol{y}_{o,*}|\boldsymbol{x}_*,\boldsymbol{\theta},\bm\Sigma_y) \notag\\
    &\times p(\boldsymbol{X},\boldsymbol{\theta},\bm\Sigma_y|\boldsymbol{Y}) {\rm d}\boldsymbol{X}{\rm d}\boldsymbol{\theta}{\rm d}\boldsymbol{\Sigma}_y,
\end{align}
where $p(\boldsymbol{y}_{o,*}|\boldsymbol{x}_*,\boldsymbol{\theta},\bm\Sigma_y)$ is a Gaussian distribution, and $p(\boldsymbol{X},\boldsymbol{\theta}, \boldsymbol{\Sigma}_y|\boldsymbol{Y})$ is given by \eqref{eq:neweq1}. 

\item Next we find the predictive distribution of the unobserved components $y_{d,*}$, $d\in{\cal S}_u$, where ${\cal S}_u$ represents the set of indices corresponding to the unobserved outputs. This is done using the GPs associated with the respective components of $\boldsymbol{y}_{u,*}$, which leads to the predictive distribution
\begin{align}
\label{eq:preddis}
    p(y_{d,*}|\boldsymbol{y}_{o,*},\boldsymbol{Y})&\propto\int  p(y_{d,*}|\boldsymbol{x}_*,\boldsymbol{\theta},\bm \Sigma_y)p(\boldsymbol{x}_*|\boldsymbol{y}_{o,*},\boldsymbol{\theta},\bm\Sigma_y)\notag\\
    &\times p(\boldsymbol{X},\boldsymbol{\theta},\bm\Sigma_y|\boldsymbol{Y}) \, {\rm d}\boldsymbol{X} {\rm d}\boldsymbol{\theta}{\rm d}\boldsymbol{x}_* {\rm d}\bm\Sigma_y,
\end{align}
where  $p(\boldsymbol{x}_*|\boldsymbol{y}_{o,*},\boldsymbol{\theta},\bm\Sigma_y)$ was determined in step 1, and $p(\boldsymbol{X},\boldsymbol{\theta},\bm\Sigma_y|\boldsymbol{Y})$ is defined by \eqref{eq:neweq1}.

\item 
From \eqref{eq:preddis}, we can draw samples of each $y_{d,*}$ and use them to approximate its predictive distribution. Often, this predictive distribution is approximated by a Gaussian factorized across dimensions, i.e.,   
\begin{align}
\label{eq:prediction}
    p(\bm y_{u,*}|\boldsymbol{y}_{o,*},\boldsymbol{Y}) &= \prod_{d\in {\cal S}_u} \mathcal{N}(y_{d,*} | \mu_{d,*}, \sigma_{d,*}^2),
\end{align}
where $\mu_{d,*}$ is the posterior GP mean for dimension $d$, and $\sigma_{d,*}^2$ is the corresponding posterior variance. The variance $\sigma_{d,*}^2$ quantifies the uncertainty in each predicted component of $\bm y_{u,*}$ and it includes both aleatoric and epistemic uncertainty, where the epistemic uncertainty arises from the uncertainty of the latent variables
and the model parameters.  
\end{enumerate}

In order to expand our work on UQ to as wide a set of problems as possible, we explore the use of Random Fourier Features (RFF)--based GPs \cite{rahimi2007random}. These GPs approximate an adopted kernel function by a finite-dimensional RFF  mapping by  
\begin{align}
    \kappa(\boldsymbol{x},\boldsymbol{x}^\prime)&\approx \boldsymbol{\phi}(\boldsymbol{x})^\top\boldsymbol{\phi}(\boldsymbol{x}^\prime),
\end{align}
where $\boldsymbol{\phi}(\boldsymbol{x})\in\mathbb{R}^{J\times 1}$ is given by  
\begin{align}
\boldsymbol{\phi}(\boldsymbol{x})&=\sqrt{\frac{2}{J}}
\left[
\begin{matrix}
    \cos(\boldsymbol{\omega}_1^\top\boldsymbol{x}), 
    \sin(\boldsymbol{\omega}_1^\top\boldsymbol{x}),
    \dots,
    \sin(\boldsymbol{\omega}_{J/2}^\top\boldsymbol{x})
\end{matrix}
\right]^\top,
\end{align}
where $\boldsymbol{\omega}_j$ are random frequencies, which according to Bochner's theorem \cite{rudin2017fourier},  are drawn from the power spectral density of the chosen kernel. The functions of interest shown in Fig. \ref{Fig:GPLVM} are then modeled by
\begin{align}
f_d(\boldsymbol{x}_n)&=\boldsymbol{\phi}(\boldsymbol{x}_n)^\top\boldsymbol{\theta}_d.
\end{align}

Thus, after training the RFF-based GPs, the log-likelihood used to optimize $\bm x_*$ becomes   
\begin{align}
    {\cal L}
    &=\sum_{d \in \mathcal{S}_o} \log \mathcal{N}(y_{o,*,d} | \bm{\phi}(\boldsymbol{x}_*)^\top \widehat{\bm{\theta}}_d, \widehat{\sigma}_{o,*,d}^2) + \log p(\bm x_*),
\end{align}
where $\widehat{\bm{\theta}}_d$ and $\widehat{\sigma}_{o,*,d}^2$ represent estimated values from the training data.

\section{Uncertainty quantification: The theory}
\label{sec:theory}
Given the above description of GPLVMs, we first formulate the problem.  We are given a training dataset \( \mathcal{D} = \{\bm y_n\}_{n=1}^{N} \), which is used to construct a GPLVM. The construction involves estimating the latent variables \( \bm X \), the parameters of the GP models, \( \bm \Theta=[\bm\theta_1,\dots, \bm\theta_d]\), and \(\bm \Sigma_y\). The estimates of \( \bm X \),  \( \bm \Theta \), and \(\bm \Sigma_y\) can be samples from $p(\boldsymbol{X},\boldsymbol{\theta}, \bm \Sigma_y|\boldsymbol{Y})$, which we denote by \( \bm X^{(m)} \),  \( \bm \Theta^{(m)} \), and \(\bm \Sigma_y^{(m)}\), $m=1, \dots, M$.
After training, a test observation \( \bm y_* \) is provided, where some elements are missing. The missing variables are predicted using Gaussian predictive distributions constructed by the GPs that correspond to these variables. {\em Our goal is to quantify the epistemic and aleatoric uncertainties associated with these predictions}. 

As already pointed out, we use the law of total variance and decompose the total predictive variance of \( y_{d,*} \) by 
\begin{align}
    \text{Var}[y_{d,*}] &= \underbrace{\mathbb{E}_{p(\bm{x}_* | \bm{y}_{o,*}, \bm{Y})} 
    \big[ \text{Var}_{p(y_{d,*} | \bm{x}_*, \bm{\theta}_d)} [y_{d,*}] \big]}_{\text{Aleatoric Uncertainty}} \notag \\
    &\quad + \underbrace{\text{Var}_{p(\bm{x}_* | \bm{y}_{o,*}, \bm{Y})} 
    \big[ \mathbb{E}_{p(y_{d,*} | \bm{x}_*, \bm{\theta}_d)} [y_{d,*}] \big]}_{\text{Epistemic Uncertainty}}.
\end{align}
We focus first on the epistemic uncertainty when making predictions of $y_{d,*}$. This uncertainty comes from three main sources
\begin{enumerate}
    \item Uncertainty in training latent variables \( \bm{X} \): Recall that \( \bm{X} \) was inferred from data and that it follows a posterior distribution \( p(\bm{X} | \bm{Y}) \), which affects both \( \bm{\theta}_d \) and \( \bm{x}_* \).
    \item {Uncertainty in test latent variable \( \bm{x}_* \)}: The posterior \( p(\bm{x}_* | \bm{y}_{o,*}, \bm{Y}) \) also introduces variability in predictions.
    \item {Uncertainty in function parameters \( \bm{\theta}_d \)}: Since \( \bm{\theta}_d \) is also inferred from data, its posterior \( p(\bm{\theta}_d | \bm{X}, \bm{Y}) \) also contributes to epistemic uncertainty.
\end{enumerate}

The predicted value of $y_{d,*}$ has a mean that is linear in \( \bm{\theta}_d \), i.e.,
\begin{align}
    \mathbb{E}_{p(y_{d,*} | \bm{x}_*, \bm{\theta}_d)} [y_{d,*}] = \bm{\phi}(\bm{x}_*)^\top \mathbb{E}[\bm{\theta}_d | \bm{X}, \bm{Y}].
\end{align}
Thus, the epistemic uncertainty simplifies to
\begin{align}
    \sigma_{\text{epistemic}, d,*}^2 = \text{Var}_{p(\bm{x}_* | \bm{y}_{o,*}, \bm{Y})} 
    \big[ \bm{\phi}(\bm{x}_*)^\top \mathbb{E}[\bm{\theta}_d | \bm{X}, \bm{Y}] \big].
\end{align}
Now we use the law of total variance, and we decompose the epistemic uncertainty into contributions from the training latent variables \( \bm{X} \) and test latent variable \( \bm{x}_* \),

We incorporate the uncertainty in \( \bm{X} \) by applying the law of total variance again. We write 
\begin{align}
    &\sigma_{\text{epistemic}, d,*}^2\notag\\
    &= 
    \mathbb{E}_{p(\bm{X} | \bm{Y})} \Big[
    \text{Var}_{p(\bm{x}_* | \bm{y}_{o,*}, \bm{X}, \bm{Y})} 
    \big[ \bm{\phi}(\bm{x}_*)^\top \mathbb{E}[\bm{\theta}_d | \bm{X}, \bm{Y}] \big] \Big] \notag \\
    &\quad + \text{Var}_{p(\bm{X} | \bm{Y})} \Big[
    \mathbb{E}_{p(\bm{x}_* | \bm{y}_{o,*}, \bm{X}, \bm{Y})} 
    \big[ \bm{\phi}(\bm{x}_*)^\top \mathbb{E}[\bm{\theta}_d | \bm{X}, \bm{Y}] \big] \Big],
\end{align}
where the first term accounts for uncertainty due to test input variability, reflecting how variations in \( \bm{x}_* \) variations influence the prediction, and the second term accounts for uncertainty from inferred latent variables and represents the impact of different inferred values of  \( \bm{X} \).

We recall that 
\begin{align}
    \mathbb{E}_{p(y_{d,*} | \bm{x}_*, \bm{\theta}_d)} [y_{d,*}] = \bm{\phi}(\bm{x}_*)^\top \bm{\theta}_d,
\end{align}
and write
\begin{align}
    &\text{Var}_{p(\bm{x}_* | \bm{y}_{o,*}, \bm{X}, \bm{Y})} 
    \big[ \bm{\phi}(\bm{x}_*)^\top \mathbb{E}[\bm{\theta}_d | \bm{X}, \bm{Y}] \big] \notag\\
    &\quad = 
    \mathbb{E}_{p(\bm{x}_* | \bm{y}_{o,*}, \bm{X}, \bm{Y})} 
    \big[ \bm{\phi}(\bm{x}_*)^\top \text{Cov}[\bm{\theta}_d | \bm{X}, \bm{Y}] \bm{\phi}(\bm{x}_*) \big] \notag \\
    &\quad + \text{Var}_{p(\bm{x}_* | \bm{y}_{o,*}, \bm{X}, \bm{Y})} 
    \big[ \bm{\phi}(\bm{x}_*)^\top \mathbb{E}[\bm{\theta}_d | \bm{X}, \bm{Y}] \big].
\end{align}
Next, we take the expectation over \( p(\bm{X} | \bm{Y}) \) and obtain
\begin{align}
    &\sigma_{\text{epistemic}, d,*}^2 = 
    \mathbb{E}_{p(\bm{X} | \bm{Y})} \Bigg[
    \bm{\mu}_\phi^\top \text{Cov}[\bm{\theta}_d | \bm{X}, \bm{Y}] \bm{\mu}_\phi\notag\\
    &\quad+ \text{Tr} \big( \text{Cov}_{p(\bm{x}_* | \bm{y}_{o,*}, \bm{X}, \bm{Y})} [\bm{\phi}(\bm{x}_*)] \text{Cov}[\bm{\theta}_d | \bm{X}, \bm{Y}] \big) \Bigg] \notag \\
    \label{eq:ep}
    &\quad + \text{Tr} \big( \text{Cov}_{p(\bm{x}_* | \bm{y}_{o,*}, \bm{Y})} [\mathbb{E}_{p(y_{d,*} | \bm{x}_*, \bm{\theta}_d)} [y_{d,*}]] \big),
\end{align}
where
\begin{align}
    \bm{\mu}_\phi&=\mathbb{E}_{p(\bm{x}_* | \bm{y}_{o,*}, \bm{X}, \bm{Y})} \big[\bm \phi(\bm x_*) \big].
\end{align}
We observe that 
\begin{description}
    \item[(a)] the uncertainty of the function parameters is  quantified by \(p(\bm{\theta}_d | \bm{X}, \bm{Y}) \), more specifically by \( \text{Cov}[\bm{\theta}_d | \bm{X}, \bm{Y}] \).  We note it affects the predictions through both expectation and variance terms,
    \item[(b)] the uncertainty due to $\bm x_*$ 
is quantified by \( p(\bm{x}_* | \bm{y}_{o,*}, \bm{X}, \bm{Y}) \) and is tracked via \( \text{Var}_{p(\bm{x}_* | \bm{y}_{o,*}, \bm{X}, \bm{Y})} [\cdot] \), and 
\item[(c)] the effect of the uncertainty  on the predictions due to  \( \bm{X} \) and quantified by \( p(\bm{X} | \bm{Y}) \) appears in \( \text{Var}_{p(\bm{X} | \bm{Y})} [\cdot] \).
\end{description}  

Now we turn our attention to the aleatoric uncertainty. In our GPLVM framework, we assume that
\begin{align}
    y_{d,*} \sim \mathcal{N}(\bm{\phi}(\bm{x}_*)^\top \bm{\theta}_d, \sigma_{d}^2),
\end{align}
and, thus, 
\begin{align}
    \text{Var}_{p(y_{d,*} | \bm{x}_*, \bm{\theta}_d)} [y_{d,*}] = \sigma_{d}^2.
\end{align}
We note that since $\bm x_*$ is random,  $\sigma_d^2$ is also random. We take the expectation over \( p(\bm{x}_* | \bm{y}_{o,*}, \bm{Y}) \) and write
\begin{align}
    \sigma_{\text{aleatoric}, d*}^2 = \mathbb{E}_{p(\bm{x}_* | \bm{y}_{o,*}, \bm{Y})} [\sigma_{d}^2].
\end{align}
From training, the noise variance \( \sigma_d^2 \) is inferred from the posterior
\begin{align}
    p(\sigma_d^2 | \bm{Y}) = \int p(\sigma_d^2 | \bm{X}, \bm{Y}) p(\bm{X} | \bm{Y}) d\bm{X}.
\end{align}
Thus, the full expression for aleatoric uncertainty is
\begin{align}
\label{eq:al}
    \sigma_{\text{aleatoric}, d,*}^2 = \mathbb{E}_{p(\bm{X} | \bm{Y})} \mathbb{E}_{p(\bm{x}_* | \bm{y}_{o,*}, \bm{X}, \bm{Y})} [\sigma_{d}^2].
\end{align}

\section{Uncertainty quantification: The methods}
\label{sec:methods}

In this section we discuss the actual computation of the epistemic and aleatoric uncertainties. We propose an approximation based on a Monte Carlo sampling method. We compute the epistemic uncertainty defined by \eqref{eq:ep} using the following approach:
\begin{itemize}
    \item Draw \( M \) samples \( \bm{X}^{(m)} \sim p(\bm{X} | \bm{Y}) \).
    \item For each \( \bm{X}^{(m)} \), draw \( L \) samples $$\bm{x}_*^{(m,\ell)} \sim p(\bm{x}_* | \bm{y}_{o,*}, \bm{X}^{(m)}, \bm{Y}).$$
\end{itemize}
The Monte Carlo estimate is then
\begin{align}
\label{eq:eu}
    &\sigma_{\text{epistemic}, d,*}^2 \approx \frac{1}{M} \sum_{m=1}^{M} \Bigg[
    \bm{\mu}_\phi^\top \text{Cov}[\bm{\theta}_d | \bm{X}^{(m)}, \bm{Y}] \bm{\mu}_\phi \notag\\
    &\quad+ \text{Tr} \big( \text{Cov}_{p(\bm{x}_* | \bm{y}_{o,*}, \bm{X}^{(m)}, \bm{Y})} [\bm{\phi}(\bm{x}_*)] \text{Cov}[\bm{\theta}_d | \bm{X}^{(m)}, \bm{Y}] \big) \Bigg] \notag \\
    &\quad + \frac{1}{M} \sum_{m=1}^{M} \text{Tr} \bigg( \frac{1}{L} \sum_{\ell=1}^{L} \big( \bm{\phi}(\bm{x}_*^{(m,\ell)})^\top \mathbb{E}[\bm{\theta}_d | \bm{X}^{(m)}, \bm{Y}] \big)^2 \bigg).
\end{align}

The aleatoric uncertainty from \eqref{eq:al} is approximated by
\begin{align}
\label{eq:au}
    \sigma_{\text{aleatoric}, d,*}^2 \approx \frac{1}{M} \sum_{m=1}^{M} \Bigg[ 
    \frac{1}{L} \sum_{\ell=1}^{L} \sigma_d^{2(m)} 
    \Bigg].
\end{align}

\section{Uncertainty quantification: The insights from experiments}
\label{sec:insights}

In this section, we describe our experiments and present insights gained from them. We generated \( N = 1{,}000 \) four-dimensional observations \( \bm{y}_n \), with 800 used for training and the remaining 200 reserved for testing.
Each observed vector \(\bm{y}_n\) was obtained by first drawing a latent variable \(\bm{x}_n\in\mathbb{R}^2\) according to 
$\bm{x}_n \sim \mathcal{N}(\bm{0}, \sigma^2 \bm{I})$,
and then applying four different functions to obtain \( y_{d,n} \) for \( d = 1, 2, 3, 4 \). More specifically, the observations were generated as follows:
\begin{enumerate}
    \item linear function (baseline):
    \begin{equation}
    y_{1,n} = \bm{w}_1^\top \bm{x}_n + \epsilon_{1,n}, \quad \epsilon_{1,n} \sim \mathcal{N}(0, \sigma_\epsilon^2),
\end{equation}
 \item nonlinear squared function:
\begin{equation}
    y_{2,n} = (\bm{w}_2^\top \bm{x}_n)^2 + \epsilon_{2,n}, \quad \epsilon_{2,n} \sim \mathcal{N}(0, \sigma_\epsilon^2),
\end{equation}
\item periodic function:
\begin{equation}
    y_{3,n} = \sin(\bm{w}_3^\top \bm{x}_n) + \epsilon_{3,n}, \quad \epsilon_{3,n} \sim \mathcal{N}(0, \sigma_\epsilon^2),
\end{equation}
\item discontinuous step function:
\begin{equation}
    y_{4,n} = \begin{cases} 
        1 + \epsilon_{4,n}, & \text{if } \bm{w}_4^\top \bm{x}_n > 0, \\
        -1 + \epsilon_{4,n}, & \text{otherwise}, 
    \end{cases}, \epsilon_{4,n} \sim \mathcal{N}(0, \sigma_\epsilon^2),
\end{equation}
\end{enumerate}
where the variance of the noise in all the outputs was $\sigma_\epsilon^2=1$, and the weights $\bm w_d\in\mathbb{R}^2$ were drawn independently from ${\cal N}(\bm 0,\sigma_w^2 \bm I_2)$, with  $\sigma_w^2=1.$ 

\begin{figure}[t]
    \centering
    \includegraphics[width=\textwidth]{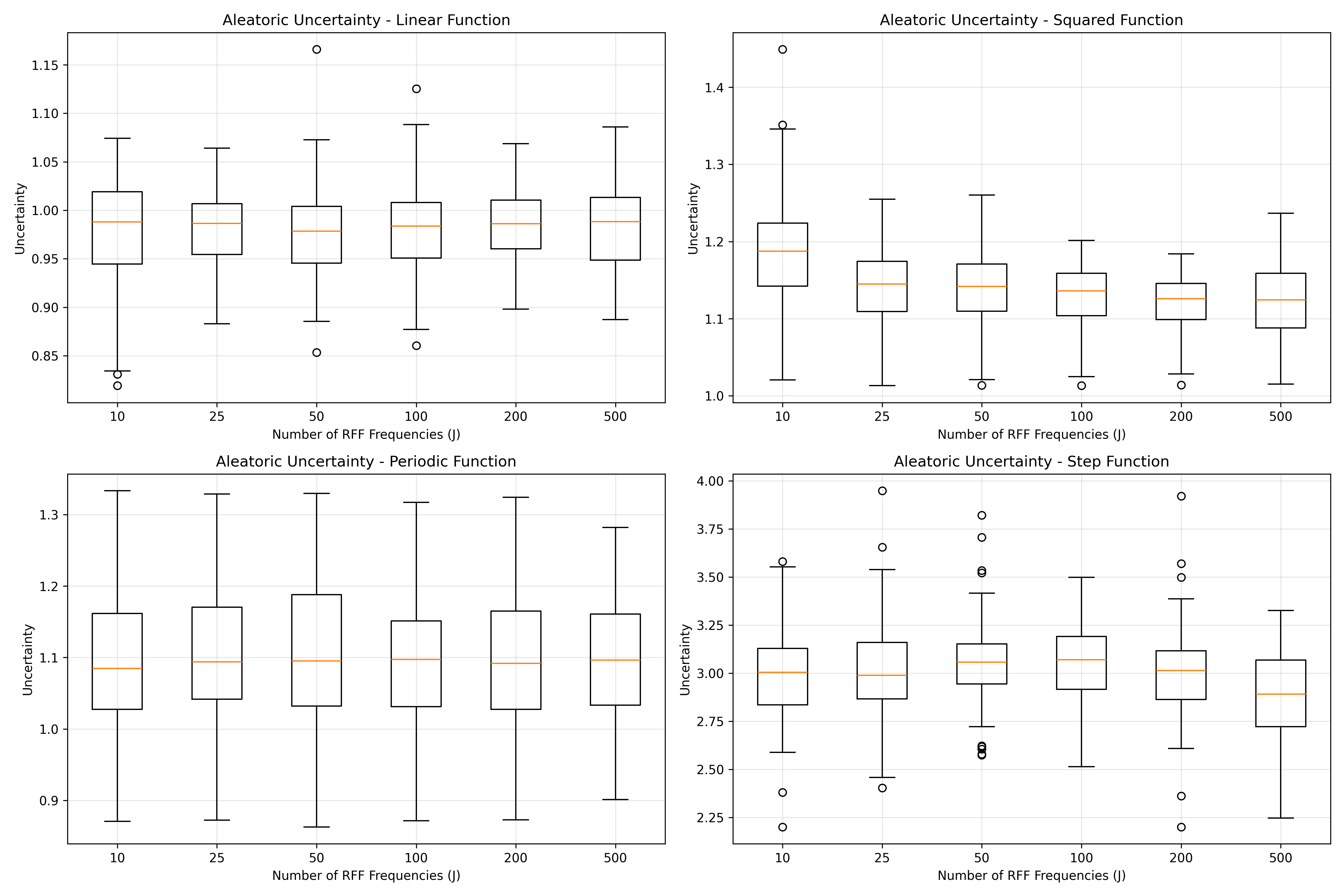}
    \caption{The estimated aleatoric uncertainties in each of  the four predicted outputs of $\bm y_*$ for various values of $J$. It was assumed that one of the outputs was missing while the remaining ones were observed.}
    \label{fig:first}
\end{figure}

During training, we drew $M=100$ samples from $p(\bm X|\bm Y)$ using the variational distribution of $\bm X$. For each test sample, \(\bm{y}_{o,*}\) and \( \bm{X}^{(m)} \), we generated $L=100$ samples of $\bm x_*$ according to \( \bm{x}_*^{(m,\ell)} \sim p(\bm{x}_* | \bm{y}_{o,*}, \bm{X}^{(m)}, \bm{Y}) \). We used the generated training samples $\bm X^{(m)}$ to compute the epistemic and aleatoric uncertainties defined by \eqref{eq:eu} and \eqref{eq:au}, respectively. We conducted experiments where the number of drawn frequencies took values from the set $J\in\{10, 25, 50, 100, 200, 500\}$. All the frequencies were obtained by sampling from the radial-basis-function kernel.   

\begin{figure}[t]
    \centering
    \includegraphics[width=\textwidth]{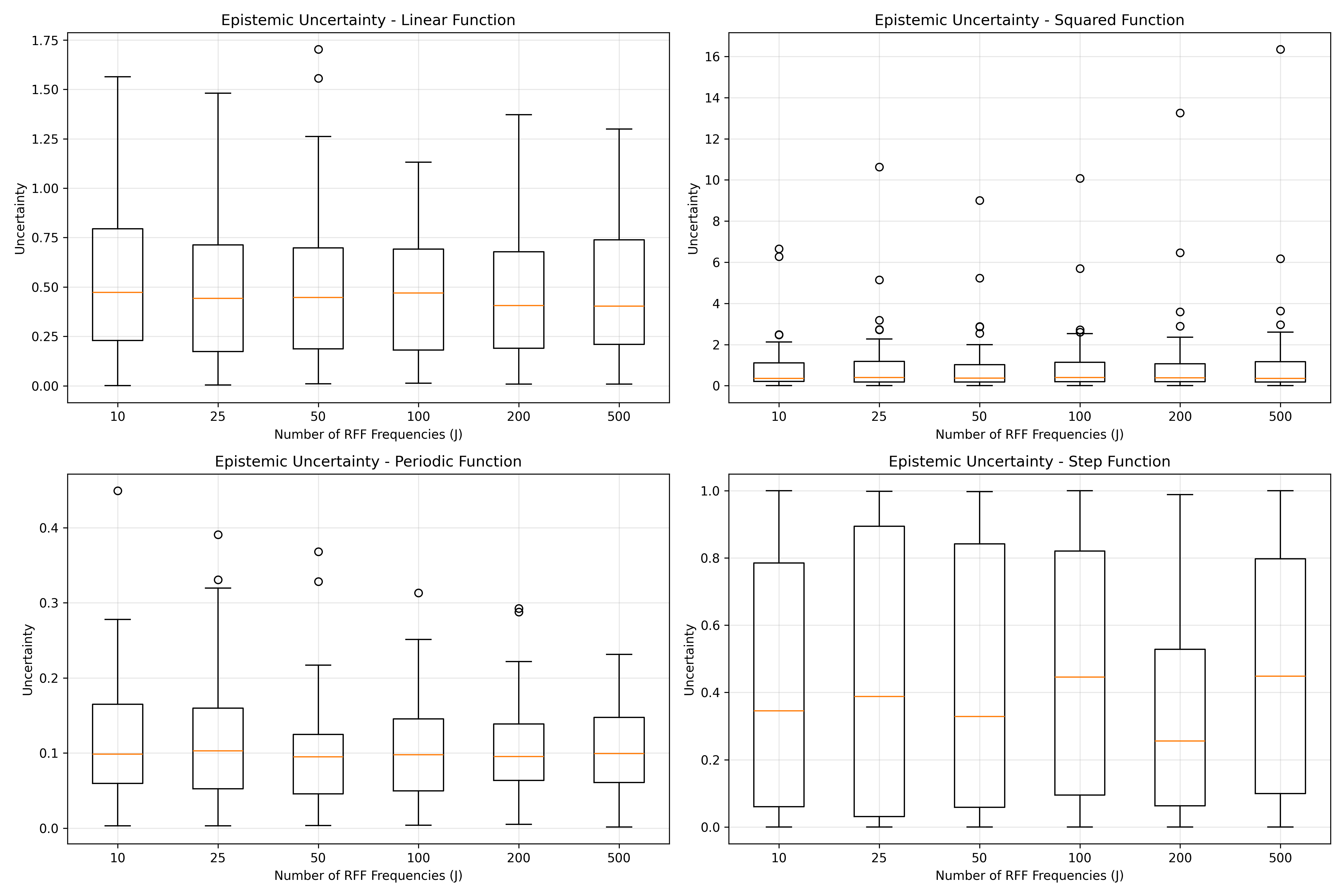}
    \caption{The estimated epistemic uncertainties in each of  the four predicted outputs of $\bm y_*$ for various values of $J$. It was assumed that one of the outputs was missing while the remaining ones were observed.}
    \label{fig:second}
\end{figure}

During evaluation, we computed the aleatoric and epistemic uncertainties as described in the previous section. In each trial, we assumed that one of the output variables, \( y_{d,*} \), was missing and had to be predicted using the remaining variables.  For example, to predict \( y_{1,*} \), we used the observed values \( y_{2,*} \), \( y_{3,*} \), and \( y_{4,*} \) to estimate the latent variable \( \bm{x}_* \), which was then used to infer \( y_{1,*} \). The boxplots in Figs.~\ref{fig:first} and \ref{fig:second} summarize average aleatoric and epistemic variances computed on 200 test samples per trial, following training. Each boxplot is constructed from 50 such averages obtained across independent simulations. The circles denote outlier trials. The plots illustrate the consistency of uncertainty estimates and the variability across test instances. 




We observe that the aleatoric uncertainty of \( y_{1,*} \), which corresponds to the linear function, was estimated accurately and did not vary across values of \( J \). In contrast, the aleatoric uncertainty for \( y_{4,*} \), which corresponds to the step function, was significantly overestimated due to the limited ability of GPs to model discontinuities. The epistemic uncertainty reached its highest value for \( y_{4,*} \), which reveals the model's lack of confidence in regions with abrupt changes, and its lowest value for \( y_{3,*} \), the periodic function output, which is smooth and more suitable for GP models. 

\section{Conclusions}
\label{sec:conc}
In this paper, we presented a systematic framework for UQ in probabilistic ML models with focus on GPLVMs. We provided a theoretical formulation for decomposing predictive uncertainty into epistemic and aleatoric components and introduced a Monte Carlo sampling-based approach for their estimation. Using RFF-based GPs, we demonstrated a scalable and efficient method to approximate predictive distributions. Our experimental results provided insight into the impact of uncertainty estimation on model reliability. The computation of uncertainties relied on Monte Carlo simulations, which in turn introduced additional sources of uncertainty. This was not accounted for in this paper and is left for future work.

\newpage

\bibliographystyle{apalike}
\bibliography{petar-refs.bib}

\end{document}